\documentclass[conference]{IEEEtran}
\IEEEoverridecommandlockouts
\usepackage{cite}
\usepackage{amsmath,amssymb,amsfonts}
\usepackage{algorithmic}
\usepackage{textcomp}
\usepackage{xcolor}
\usepackage{graphicx}
\usepackage[ruled,linesnumbered,noresetcount]{algorithm2e}
\def\BibTeX{{\rm B\kern-.05em{\sc i\kern-.025em b}\kern-.08em
    T\kern-.1667em\lower.7ex\hbox{E}\kern-.125emX}}
\begin{document}

\title{Two-stage Fall Events Classification with Human Skeleton Data\\
}

\author{\IEEEauthorblockN{Leiyu Xie}
\IEEEauthorblockA{\textit{Intelligent Sensing and Communications } \\
\textit{Research Group, Newcastle University}\\
Newcastle upon Tyne, UK\\
l.xie6@newcastle.ac.uk}
\and
\IEEEauthorblockN{Yang Sun}
\IEEEauthorblockA{\textit{Big Data Institute } \\
\textit{University of Oxford}\\
Oxford, UK\\
yang.sun@bdi.ox.ac.uk}
\and
\IEEEauthorblockN{Jonathon A. Chambers}
\IEEEauthorblockA{\textit{School of Engineering} \\
\textit{University of Leicester}\\
Leicester, UK\\
jonathon.chambers@leicester.ac.uk}
\and
\IEEEauthorblockN{\qquad\qquad\qquad\qquad\qquad\qquad\qquad}
\IEEEauthorblockA{} \\
\textit{}
\and
\IEEEauthorblockN{ Syed Mohsen Naqvi}
\IEEEauthorblockA{\textit{Intelligent Sensing and Communications } \\
\textit{Research Group, Newcastle University}\\
Newcastle upon Tyne, UK\\
mohsen.naqvi@newcastle.ac.uk}
\and
\IEEEauthorblockN{\qquad\qquad\qquad\qquad}
\IEEEauthorblockA{} \\
\textit{}
}

\maketitle

\begin{abstract}
Fall detection and classification become an imperative problem for healthcare applications particularity with the increasingly ageing population. Currently, most of the fall classification algorithms provide binary fall or no-fall classification. For better healthcare, it is thus not enough to do binary fall classification but to extend it to multiple fall events classification. In this work, we utilize the privacy mitigating human skeleton data for multiple fall events classification. The skeleton features are extracted from the original RGB images to not only mitigate the personal privacy, but also to reduce the impact of the dynamic illuminations. The proposed fall events classification method is divided into two stages. In the first stage, the model is trained to achieve the binary classification to filter out the no-fall events. Then, in the second stage, the deep neural network (DNN) model is trained to further classify the five types of fall events. In order to confirm the efficiency of the proposed method, the experiments on the UP-Fall dataset outperform the state-of-the-art.
\end{abstract}


\section{Introduction}
\label{sec:intro}
In recent decades, falls in elderly have become one of the most important problems in the increasingly ageing population \cite{yu2017computer, yu2013online}. Over 6 million falls occur each year in the world, making it the primary reason leading to the death of elderly people \cite{wang2020elderly, tsai2019implementation}. Different types of fall events lead to different injuries in the human body parts. Thus it is not enough to do the binary fall classification but also multiple fall events classification is needed for better healthcare. 

Many approaches have been used for fall events classification. However, most of them are based on wearable sensors and it is inconvenient for ageing people to wear the sensors all the time, particularly when the dementia problem is also increased in elderly people \cite{mubashir2013survey,jiang2018smart}. In this work, a video sensor will be used for the proposed method. To mitigate the privacy information, only skeleton data are extracted from the original RGB data, which can effectively mitigate the personal information and also reduce the dynamic illumination effect on the performance \cite{abobakr2017skeleton,yan2020image}. Meanwhile, using only the skeleton data extracted from the original RGB data, can also reduce the computational cost because the skeleton data size is much smaller than RGB data to be processed \cite{angelini20192d}. Moreover, different from other human activities recognition tasks, fall is one kind of dangerous activity for elderly people that seldom occurs in the real living environment. Thus the data in the real life datasets are imbalanced. 

The dramatic impact of deep learning in recent years has greatly changed the landscape, e.g. improved the performance in many relevant tasks, such as object detection and human activities recognition. In order to mitigate the privacy mitigating and imbalanced data problem mentioned above, we propose a two-stage framework which only requires the human skeleton data for multiple fall events classification. This novel approach takes advantage of the DNN for fall events classification. In the first stage, the model will be trained with binary labels to classify the fall and no-fall. In the second stage, the framework will be applied for the multiple fall events classification based on the prior knowledge of fall from the first stage. We verify the effectiveness of the proposed framework on the UP-Fall dataset \cite{martinez2019up}. 

The outline of the paper is as follows. Section 2 provides the related work about fall detection and the related datasets. Section 3 presents the proposed two-stage fall classification framework. The detailed experimental settings and framework performance are shown in Section 4. Finally, Section 5 concludes the work.

\begin{figure*}[hbtp!]
  \centering
  \includegraphics[width=12.4cm, height=7.9cm]{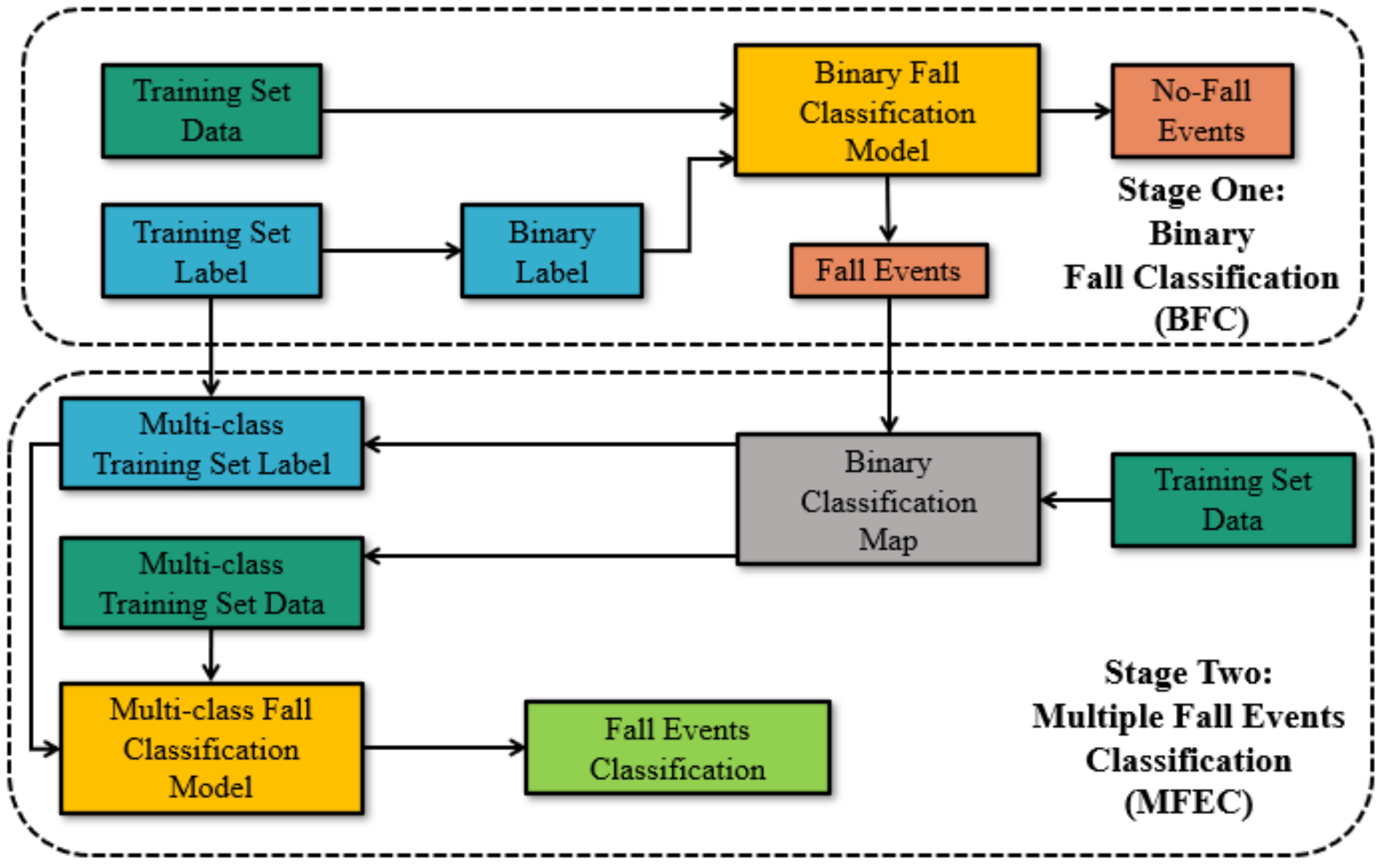}
\caption{The framework of the proposed two-stage fall events classification. Binary fall classification (BFC) provides fall and no-fall as a prior knowledge to the stage two, which yeilds the multiple fall events classification (MFEC).}
\end{figure*}

\section{Related Datasets and works}

\label{sec:relatedwork}
In this section, recently used public fall datasets and methods are briefly discussed. The Multicam dataset is one of the well-known fall detection datasets which contains 24 different scenarios with 8 cameras in the different field of views in the data recording room \cite{auvinet2010mul}. In this dataset, the first 22 scenarios contain both fall and confounding events. The remaining 2 scenarios only contain confounding events. The SDU-Fall dataset has 6 different actions with 10 subjects, one of the actions is falling to the ground \cite{ma2014depth}. UR-Fall is a relatively small scale dataset and recorded by 2 Microsoft Kinect cameras, which has 70 sequences of human activities \cite{kwolek2014human}. The UP-Fall dataset is a large scale dataset in the fall detection area which includes 5 different fall events and 6 human normal activities. In total, 17 subjects are recorded by 2 Microsoft Kinect cameras \cite{martinez2019up}. Therefore, the UP-Fall dataset is used to do the fall classification in this work.

In \cite{tsai2019implementation}, the researchers use the depth information as the input to do the binary fall detection rather than the multiple fall events classification task. By using the optical flow data, \cite{nunez2017vision} propose a transfer learning method to address the fall detection on UR-Fall and Multicam datasets. In \cite{zerrouki2018vision}, an adaptive AdaBoost classifier is used to detect fall events which occur in the UR-Fall dataset. Then, a fall detection framework based on the depth video is proposed for both SDU-Fall and UR-Fall datasets \cite{sase2018human}. A two-stream graph convolutional method, which contains both Cartesian and polar representations is proposed for skeleton data to detect binary fall \cite{liutwo}. By using the feature-extraction methods described in \cite{jeong2019human}, long short term memory (LSTM) is used to improve the accuracy of the fall-detection system. Recently, with the UP-Fall dataset, five widely-used classifiers: K-nearest neighbour (KNN), support vector machine (SVM), multilayer perceptron (MLP), random forest (RF) and Adaboost have been applied for fall detection on skeleton data achieving promising performance \cite{ramirez2021fall}. Therefore, the proposed method is compared with \cite{ramirez2021fall} in this work.

\section{Proposed Method}

Most of the fall detection methods are for binary fall classification. However, there are many kinds of fall events, such as forward falling, backward falling, which will lead to different types of injuries in human body parts. Therefore, in order to provide better healthcare for elderly people, it is crucial to enable multiple fall events classification rather than conventional binary fall classification. Therefore, we design a two-stage fall events classification framework to address the multiple fall events classification problem with privacy mitigating human skeleton data.

\begin{algorithm}[h]

\SetKwInOut{Input}{Input}\SetKwInOut{Output}{Output}

\Input{Training data $D$, original labels $L$, binary labels $L_{bin}$, binary model epoch $T_{a}^{max}$, multi-class model epoch $T_{b}^{max}$}

\BlankLine

\textbf{Initialize} Binary model $M_{a}$,\\
\qquad \qquad Multi-class classification model $M_{b}$ \\
\For{$T_{a}\leftarrow 1,2,3,...,T_{a}^{max}$}{
\textbf{Based on} $D$ and $L_{bin}$,\\
\textbf{Train} $M_{a}$ by using equation (1) ;\\
}
After $M_{a}$ is trained:\\
\textbf{Obtain} $Q_{bin} \leftarrow M_{a},D$\\
\qquad // Generate binary classification map \;

\textbf{Obtain} $D_{multi} \leftarrow D, Q_{bin}$\\
\qquad // Generate multi-class falls data \;

\textbf{Obtain} $L_{multi} \leftarrow L, Q_{bin}$\\
\qquad // Generate multi-class falls labels \;


\For{$T_{b}\leftarrow 1,2,3,...,T_{b}^{max}$}{
\textbf{Based on} $D_{multi}$ and $L_{multi}$,\\
\textbf{Train} $M_{b}$ by using equation (2) ;\\
}
\Output{Trained binary classification model $M_{a}$; Trained multi-class classification model $M_{b}$}

\caption{Two-stage fall events classification}\label{algo_disjdecomp}

\end{algorithm}
\subsection{First Stage: Binary Fall Classification (BFC)}

Due to the imbalanced data problem, the fall events classification performance will be negatively affected when trained with the normal activities data. Hence, in the BFC stage, one DNN model is trained with all available training samples in the dataset. To categorize the events into two types: fall and no-fall, processed binary labels are used as training target. The aim of the BFC is to filter out the no-fall events for eliminating the imbalanced data problem. Fall events can lead to serious danger to the health and life of elderly people, thus it is important for the model trained in the BFC stage that it can find most of the fall events and the recall measure is high.

\begin{table}[htbp!]
\centering
\caption{Comparison between the DNN and the RF \cite{ramirez2021fall} in terms of BFC. The DNN is better choice for the proposed work.}
\begin{tabular}{|c|c|c|}
\hline
Method & \begin{tabular}[c]{@{}c@{}}No. of Falls \\ Detected / Ground Truth\end{tabular} & Recall  \\ \hline
DNN    & 1788 / 1803 & 0.96   \\ \hline
RF\cite{ramirez2021fall}   & 1605 / 1803 & 0.86   \\ \hline
\end{tabular}
\end{table}

Moreover, we compare the recall measure for the trained DNN and the trained RF, because the RF achieves the best performance in \cite{ramirez2021fall}. The corresponding results are shown in Table 1. The comparison results confirm that the DNN is the better choice to select the inner model of the proposed method.

Since it is a binary classification in the BFC stage. Therefore we used binary cross entropy shown in equation (1) as the loss function. In order to overcome the information loss in the network layers, we use multiple weighted outputs in different layers as sub-outputs to reuse the information. 
\begin{equation}
loss_{a}=-\sum^{3}_{i=1}\omega_{i}(y_{i} \cdot log(\widehat{y}_{i})+(1-y_{i})\cdot log(1-\widehat{y}_{i}))
\end{equation}
where $\omega_{i}$ represents the loss weight of $i$th sub-output. $y_{i}, \widehat{y}_{i}$ indicate the ground truth and estimation for samples in the $i$th sub-output, respectively. $loss_{a}$ represents the loss function for the BFC model. The weight ratio of the sub-outputs is 1:1:2. Moreover, in the the DNN model, Sigmoid is used as the activation function of the output layer.

\subsection{Second Stage: Multiple Fall Events Classification (MFEC)}

After the DNN model in the BFC stage is trained, we obtain a subset of training data, which only contains the fall events. Then, the second DNN model is trained with the new training subset. Different from the BFC stage, the labels of the subset are the types of fall events. The second model is specifically trained to address the MFEC problem. 

Since the performance of the proposed method mostly depends on the BFC stage performance. Therefore the errors in the BFC stage would be maximally prevented. In order to reduce the classification errors, two thresholds $m$ and $n$ are used for two types of misclassifications, i.e. false positive and false negative.

\begin{equation}
loss_{b}=-\sum^{3}_{i=1}\omega_{i}(y_{i}\cdot log(\widehat{y}_{i}))
\end{equation}

The classification task is for multiple events, therefore the sparse categorical cross entropy loss function is used in the MFEC, as shown in equation (2). Weighted sub-outputs are also applied. Final loss function of MFEC is the combination of the sub-output losses and the weight ratio of the sub-outputs is 1:1:2. Sigmoid will be also used as the activation function of the output layer in the end of the model.

Once both models in the proposed method are trained. In the testing stage, the testing sample is fed into the binary model in BFC firstly, if the sample is classified as the fall-event, consequentially, it is fed into the multi-class fall classification model in MFEC to obtain the specific fall event category. The proposed method is presented in Algorithm 1.

\section{Experimental Results}
\subsection{Dataset and Pre-processing}
There are 17 subjects recorded in the UP-Fall dataset and 11 types of human activities are included, which have 5 different types of fall events as shown in Table 2. The remainder of them are normal human activities. In order to enlarge the size of the dataset, each subject's activities are recorded for 3 trials. 

\begin{table}[h]
\centering
\caption{Descriptions and abbreviations of fall events in the UP-Fall dataset.}
\begin{tabular}{c|c}
\hline
HF & Hands forward Falling  \\ \hline
KF & Knees forward Falling  \\ \hline
BF & Backward Falling      \\ \hline
SF & Sideways Falling      \\ \hline
SDF & Sit Down Falling      \\ \hline
\end{tabular}
\end{table}

In the UP-Fall dataset, there are some blank frames, with no subject. Therefore, those blank frames are removed in the pre-processing step. Then, the AlphaPose is used to extract the human skeleton data from the RGB image data \cite{fang2017rmpe}. Each skeleton dataset contains 17 key points for one subject and each key point contains 2D coordinates and confidence score. Since the UP-Fall dataset is recorded in the lab which has one glass wall. The people walking in the corridors and the glass reflection of the subjects are also recorded in the data. These non-experimental subjects' skeletons can provide negative impact to the model performance. To overcome this problem, we use a distance score to filter the noisy skeleton data. The distance between subject and camera has the positive correlation with the distance score. Only the skeleton data in all frames with the largest distance score is kept and the noisy skeleton data is removed. 

Since the UP-Fall dataset annotations are manual and semi-automatic, we introduce confident learning \cite{northcutt2021confident} to achieve the annotation cleaning and remove the samples with high probability being mislabelled. Finally, there are 217,405 groups of skeleton data in total. We split the whole dataset into two sub-groups, i.e. the training set and the testing set. There are 152,183 groups in the training set and 65,222 groups in the testing set. In the MFEC stage, for the multiple fall events classification model, there are 4,417 groups for the training set and 1,894 groups for the testing set. 

\subsection{Parameters Setting}
There are 9 dense layers in the proposed DNN model. The ReLU is used as the model inner activation function and batch normalization is also used. The concatenation is applied in the proposed DNN model. In both stages of the proposed model, Adam is applied as the optimizer for data training.

In the training stage, we set the learning rate as $0.0001$. The batch size of the BFC model is 1024 and for the second training stage is 32. The training epoch of the BFC model is 300, and for the multiple fall events classification model in the MFEC stage is 600. In order to prevent the error classification results from the BFC stage, which can affect the performance in MFEC stage. We set the threshold $m$ as 0.03 and $n$ as 0.02 for misclassification correction. The experiments are conducted on a work station with 4 GeForce GTX 1080Ti GPUs, and 16GB of RAM.

\subsection{Results and Discussion}
In Table 1, we already show that the DNN outperforms RF in terms of the BFC. We only present the results of proposed multiple fall events classification (MFEC) in this section. Moreover, we compare the five classifiers used in \cite{ramirez2021fall} for the MFEC in Fig. 2, which show that the RF is the best classifier in terms of the performance.
\begin{figure}[h]
\centering
\includegraphics[width=8.6cm, height=5.4cm]{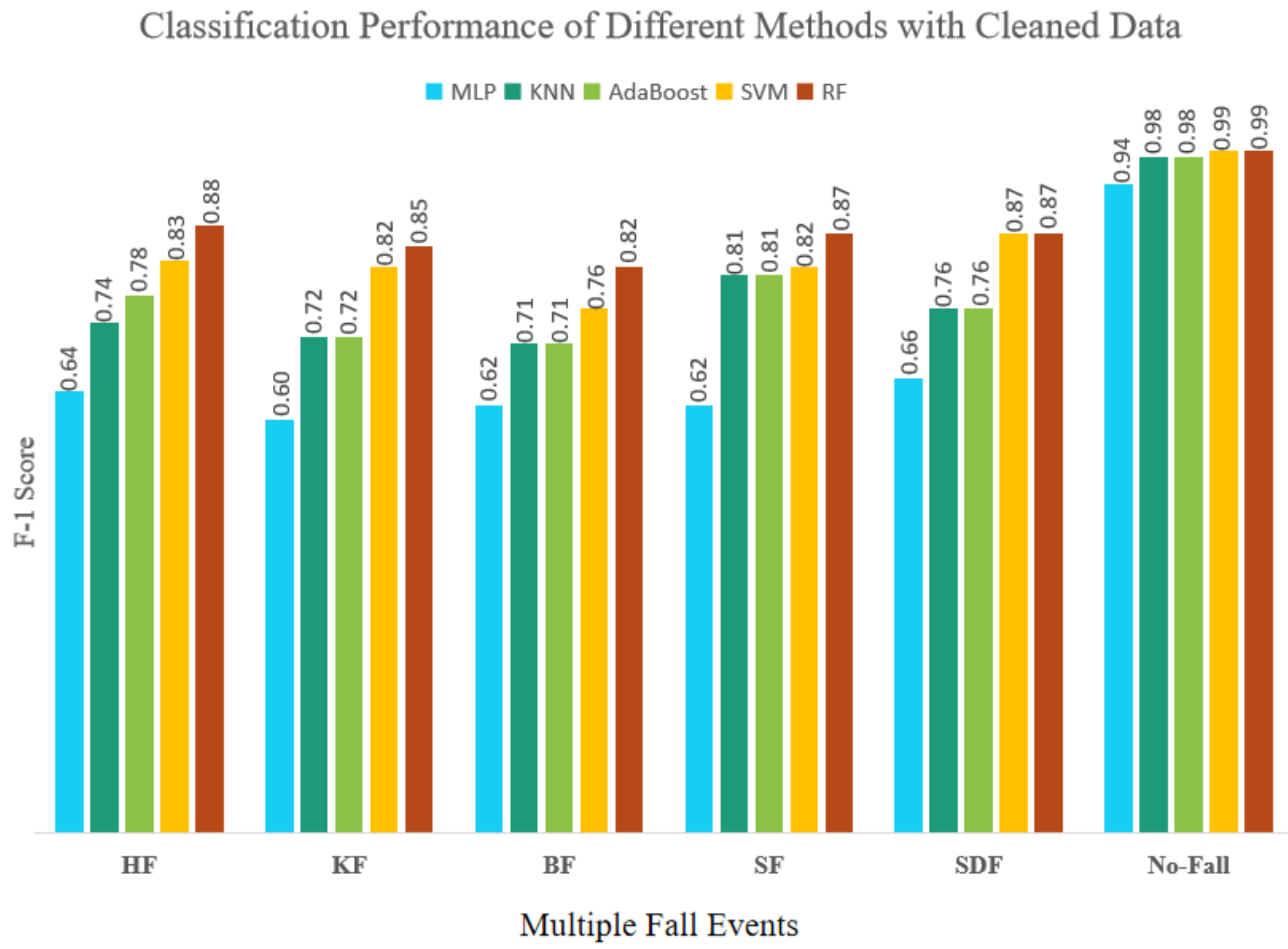}
\caption{The performance of classification by using the cleaned UP-Fall dataset. The RF achieves the best performance in all event classifiers.} 
\end{figure}

Firstly, we compare the single RF with the single DNN in Table 3. It can be observed that the single DNN can achieve better classification performance for BF, SF and SDF. With the proposed two-stage DNN (TS-DNN), the overall multiple events classification performance is further improved. Meanwhile, no thresholds are added into the TS-DNN which is shown in Table 3 ($m$ = 0 and $n$ = 0). However, the hands falling (HF) performance is lower than the single model. We assume the reason is the misclassification of the binary fall classification model in the BFC stage. Therefore, two thresholds $m$ and $n$ will be used to re-classify the misclassification of fall and no-fall categories in the BFC stage, which could improve the fall events performance in the MFEC stage.

\begin{table}[h]
\centering
\caption{Comparison of multiple fall events classification with cleaned data to present the worst-case-performance improvement. The F1-scores are provided.}
\begin{tabular}{l|c|c|c|c|c}
\hline
Methods & HF & KF & BF & SF & SDF  \\ \hline
Single RF \cite{ramirez2021fall}   & \textbf{0.88}   & 0.85 & 0.82 & 0.87 & 0.87 \\ 
\textit{Single DNN} & \textit{0.85} & \textit{0.83} & \textit{0.83} & \textit{0.88} & \textit{0.88} \\
\textit{TS-DNN (m,n=0)} & \textit{0.84} & \textit{\textbf{0.85}} & \textit{\textbf{0.85}} & \textit{\textbf{0.89}} & \textit{\textbf{0.89}} \\ \hline
\end{tabular}
\end{table}

According to Table 4, it can be observed that after adding the thresholds, the TS-DNN shows better performance than without thresholds shown in Table 3. The proposed TS-DNN with thresholds ($m$ = 0.03 and $n$ = 0.02) further improves the F1-score in terms of HF and KF. It is confirmed that the performance of both HF and KF which include misclassification in the BFC stage are revised and improved.

\begin{table}[h]
\centering
\small\addtolength{\tabcolsep}{-1pt}
\caption{Multiple fall events classification performance by using different threshold settings with cleaned data. The F1-scores are provided.}
\begin{tabular}{l|c|c|c|c|c}
\hline
Methods & HF & KF & BF & SF & SDF  \\ \hline
\textit{TS-DNN (m,n=0.3)} & \textit{0.72} & \textit{0.76} & \textit{0.63} & \textit{0.74} & \textit{0.82} \\ 
\textit{TS-DNN (m,n=0.1)} & 0.84  & 0.86  & 0.84  & 0.87  & 0.88 \\ 
\textit{TS-DNN (m,n=0.05)} & \textit{0.85} & \textit{0.86} & \textit{0.85} & \textit{0.88} & \textit{0.89}\\ 
\textit{\textbf{TS-DNN (m,n=0.03,0.02)}}  & \textit{\textbf{0.85}} & \textit{\textbf{0.86}} & \textit{\textbf{0.85}} & \textit{\textbf{0.89}} & \textit{\textbf{0.89}} \\ 
\textit{TS-DNN (m,n=0)} & \textit{0.84} & \textit{0.85} & \textit{0.85} & \textit{0.89} & \textit{0.89} \\ \hline
\end{tabular}
\end{table}

In summary, according to the F1-score, generally, it is confirmed that the proposed two-stage framework can achieve the highest accuracy in multiple fall events classification. After the DNN-based binary classifier is introduced in the BFC stage, the data imbalanced problem is eliminated and the multi-class classifier can be better trained in the MFEC stage. Besides, adding thresholds in the BFC stage could help to improve the classification performance of the MFEC stage.

\section{Conclusion}
\label{sec:majhead}
In this paper, based on the skeleton data extracted from UP-Fall dataset, we proposed a novel two-stage DNN-based framework to achieve the multiple fall events classification. By using skeleton data, the dynamic illumination and privacy mitigating issues were addressed. Then, in the proposed BFC stage of the proposed method, both the imbalanced classification and binary fall classification are performed. Finally, the multiple fall events classification problem is performed in the MFEC stage. The experimental results confirmed the proposed method could earn higher classification accuracy compared with the state-of-the-art classifiers. 

\bibliographystyle{IEEEbib}
\bibliography{main}

\end{document}